\documentclass[sigconf]{acmart}

\usepackage{algorithm}
\usepackage{algpseudocodex}
\usepackage{graphicx}
\usepackage{textcomp}
\usepackage{xcolor}
\usepackage{enumitem}
\usepackage{setspace}
\usepackage{subcaption}
\usepackage{makecell}
\usepackage{booktabs}

\title{ScoutAttention: Efficient KV Cache Offloading via Layer-Ahead CPU Pre-computation for LLM Inference}

\author{Qiuyang Zhang}
\affiliation{
  \institution{Huazhong University of Science and Technology}
  \city{Wuhan}
  \country{China}
}
\email{qyzhang@hust.edu.cn}

\author{Kai Zhou}
\affiliation{
  \institution{Huazhong University of Science and Technology}
  \city{Wuhan}
  \country{China}
}
\email{kezzy@hust.edu.cn}

\author{Ding Tang}
\affiliation{
  \institution{Huazhong University of Science and Technology}
  \city{Wuhan}
  \country{China}
}
\email{dinger@hust.edu.cn}

\author{Kai Lu}
\authornote{Kai Lu is the corresponding author}
\affiliation{
  \institution{Huazhong University of Science and Technology}
  \city{Wuhan}
  \country{China}
}
\email{kailu@hust.edu.cn}

\author{Cheng Li}
\affiliation{
  \institution{Huawei Technologies}
  \city{Hefei}
  \country{China}
}
\email{licheng56@huawei.com}

\author{Zhenyu Yang}
\affiliation{
  \institution{Huawei Technologies}
  \city{Hefei}
  \country{China}
}
\email{yangzhenyu33@huawei.com}

\author{Peng Xu}
\affiliation{
  \institution{Research Center for High Efficiency Computing Infrastructure\\ Zhejiang Lab}
  \city{Hangzhou}
  \country{China}
}
\email{xup@zhejianglab.com}

\author{Jiguang Wan}
\affiliation{
  \institution{Huazhong University of Science and Technology}
  \city{Wuhan}
  \country{China}
}
\email{jgwan@hust.edu.cn}

\copyrightyear{2026}
\acmYear{2026}
\setcopyright{cc}
\setcctype{by-nc-nd}
\acmConference[DAC '26]{63rd ACM/IEEE Design Automation Conference}{July 26--29, 2026}{Long Beach, CA, USA}
\acmBooktitle{63rd ACM/IEEE Design Automation Conference (DAC '26), July 26--29, 2026, Long Beach, CA, USA}
\acmDOI{xxxxxxxxxxxxxxxxxxxxxxx}
\acmISBN{xxxxxxxxxxxxxxxxxxxxxxxxx}

\renewcommand{\shortauthors}{Qiuyang Zhang et al.}

\begin{document}

\begin{abstract}
Large language models encounter critical GPU memory capacity constraints during long-context inference, where KV cache memory consumption severely limits decode batch sizes. While existing research has explored offloading KV cache to DRAM, these approaches either demand frequent GPU-CPU data transfers or impose extensive CPU computation requirements, resulting in poor GPU utilization as the system waits for I/O operations or CPU processing to complete.

We propose ScoutAttention, a novel KV cache offloading framework that accelerates LLM inference through collaborative GPU-CPU attention computation.
To prevent CPU computation from bottlenecking the system, ScoutAttention introduces GPU-CPU collaborative block-wise sparse attention that significantly reduces CPU load. Unlike conventional parallel computing approaches, our framework features a novel layer-ahead CPU pre-computation algorithm, enabling the CPU to initiate attention computation one layer in advance, complemented by asynchronous periodic recall mechanisms to maintain minimal CPU compute load.
Experimental results demonstrate that ScoutAttention maintains accuracy within 2.4\% of baseline while achieving 2.1x speedup compared to existing offloading methods.
\end{abstract}

\maketitle

\section{Introduction}

In recent years, Large Language Models (LLMs) have achieved unprecedented progress across a wide range of application domains \cite{llm-survey}. Traditionally, these advances have been driven by increasing model size and training data---a paradigm known as \emph{train-time scaling} \cite{scaling-law}. However, with the diminishing returns of train-time scaling, the field is gradually shifting toward \emph{inference-time scaling} strategies \cite{inference-scaling}. Pioneered by models such as OpenAI's o1 \cite{openai-o1} and DeepSeek-R1 \cite{deepseek-r1}, inference-time scaling significantly extends output length through a ``think-before-answer'' approach, employing chain-of-thought (CoT) reasoning that enables models to perform more extensive computation during inference.

Concurrently, to fully harness the capabilities of LLMs, real-world applications increasingly depend on sophisticated context engineering techniques \cite{context-engineering}. Approaches such as Retrieval-Augmented Generation (RAG) \cite{rag} and few-shot prompting inject contextual information into prompts, resulting in longer inputs that enable LLMs to better understand user queries and produce more accurate responses.

While these two trends, extending input length through context engineering and extending output length through chain-of-thought reasoning, have significantly improved LLM performance, their convergence places substantial pressure on the decoding stage.
Specifically, the decoding is bottlenecked by GPU memory capacity.
Since decoding is memory-bound, effective batching is crucial for improving GPU utilization and throughput. However, GPU memory capacity constraints limit the maximum batch size, as longer sequences consume substantially more memory.
For example, given that a 32k-token Qwen3-32B request consumes 8GB for the KV cache, the total memory usage (including weights and activations) on an 80GB GPU limits the batch size to one, which greatly limits inference throughput.

To address the memory bottleneck, existing work attempts to offload part of the KV cache to DRAM during inference. For example, InfiniGen \cite{infinigen} offloads less important tokens to DRAM after the prefill phase, retaining only a subset of tokens on the GPU to reduce memory usage. During the decode stage, it predicts the important tokens for the next layer one-layer-ahead and prefetches them back to the GPU. Although this approach mitigates the GPU memory capacity constraint, it introduces a new I/O bottleneck.
We observe that in InfiniGen, even with prefetching, slow I/O causes the GPU to stall for 61\% of the end-to-end execution time, leading to a substantial performance degradation.

Beyond recall-based KV cache offloading methods, there also exist co-attention approaches that compute attention directly on the CPU. For example, HGCA \cite{hgca} avoids recalling KV tokens back to the GPU; instead, tokens offloaded to the CPU are processed concurrently on the CPU alongside GPU computation. Although this parallel attention strategy eliminates the I/O bottleneck in recall-based methods, it introduces a new compute bottleneck on the CPU. Given that GPU attention computation is approximately 20x faster than CPU, this parallel execution scheme still results in GPU idle time. Our experimental results reveal that with HGCA, the GPU remains idle for 57\% of its execution time while waiting for CPU attention computation to complete.

We propose \textbf{ScoutAttention}, an efficient KV cache offloading framework for LLM inference. Since CPU-side attention computation achieves higher effective throughput than PCIe data transfer, ScoutAttention adapts a co-attention strategy to eliminate slow recall I/O.
To prevent CPU computation from becoming the new performance bottleneck, ScoutAttention introduces a novel \textbf{layer-ahead CPU pre-computation} mechanism. Unlike conventional parallel execution, this approach initiates CPU computation one layer in advance, effectively hiding latency through pipelining. 
Building on this foundation, ScoutAttention implements \textbf{GPU-CPU collaborative block-wise sparse attention}, which strategically retains only the most critical KV cache blocks on the GPU while offloading the remainder to CPU memory. Our key observation is that the importance distribution of KV cache blocks exhibits strong temporal locality---adjacent decode tokens typically attend to highly overlapping sets of important blocks.
By exploiting this property, ScoutAttention limits CPU computation to only a small subset of top-k blocks not present in the GPU cache, significantly reducing computational overhead while preserving model accuracy.

Moreover, to avoid increased CPU workload caused by temporal drift in the set of important blocks, we introduce \textbf{asynchronous periodic KV cache recall}, which periodically recalls important KV blocks to the GPU to correct the drift. Unlike prior recall-based methods, periodic recall in ScoutAttention is issued only \emph{after} a layer finishes. Compared with InfiniGen's one-layer-ahead recall, our asynchronous design stretches the I/O window from a single layer to an entire decode step of one token, ensuring that the GPU never stalls waiting for I/O.

We implement ScoutAttention on SGLang \cite{sglang} and evaluate its accuracy and performance across multiple datasets. The results show that the average accuracy degradation of ScoutAttention compared with full attention is less than 2.1\%, while its decoding throughput reaches 5.1x that of full attention and 2.1x that of existing offloading methods.

\section{Background and Motivation}

\begin{figure}[tb]
  \vspace*{.5em}
  \captionsetup[subfigure]{justification=raggedright, singlelinecheck=false}
  \centering
  \begin{subfigure}{0.5\textwidth}
    \centering
    \includegraphics[width=\linewidth]{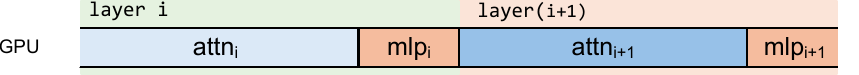}
    \caption{Full Attention: high attention computation under long-context.}
    \label{fig:attn_time_a}
    \vspace{3pt}
  \end{subfigure}
    
  \begin{subfigure}{0.5\textwidth}
    \includegraphics[width=\linewidth]{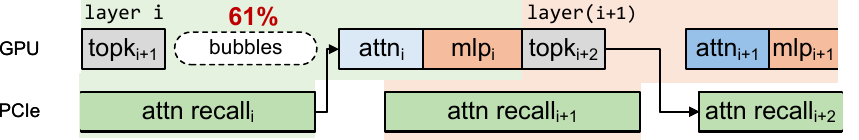}
    \caption{InfiniGen: pipeline bubbles due to slow KV cache recall.}
    \label{fig:attn_time_b}        
    \vspace{3pt}
  \end{subfigure}

  \begin{subfigure}{0.5\textwidth}
    \includegraphics[width=\linewidth]{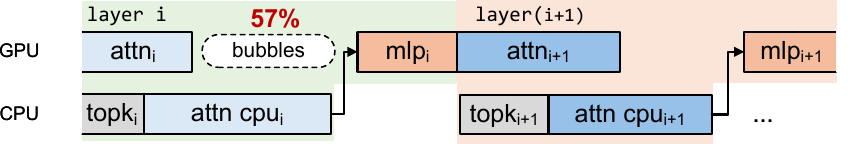}
    \caption{HGCA: pipeline bubbles due to slow CPU-side computation.}
    \label{fig:attn_time_c}        
    \vspace{3pt}
  \end{subfigure}

  \begin{subfigure}{0.5\textwidth}
    \includegraphics[width=\linewidth]{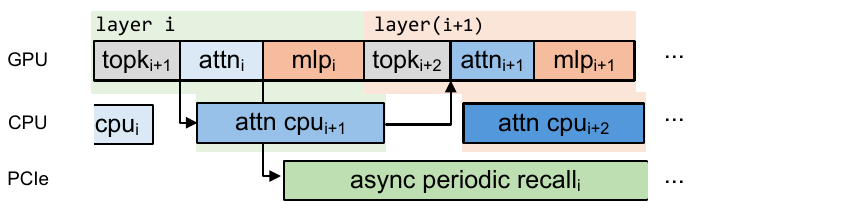}
    \caption{ScoutAttention: efficient inference pipeline through layer-ahead CPU pre-computation.}
    \label{fig:attn_time_d}
  \end{subfigure}

  \caption{Inference pipeline of different KV cache offloading methods.}
  \label{fig:pipeline}

  \vspace*{-.5em}
\end{figure}

\subsection{LLM Inference and Prefill-Decode Disaggregation}

Large language models (LLMs) implement autoregressive sequence modeling with a deep stack of Transformer blocks \cite{transformer}, each comprising multi-head attention and a feed-forward network (FFN) with residual connections.

During inference, this architecture operates in two distinct phases.
The \textbf{prefill} phase processes the entire input prefix in one pass, generating and storing the $K$ and $V$ vectors in the \textbf{KV cache}. The \textbf{decode} phase then produces tokens autoregressively, reusing the cached K/V to compute each subsequent token.
Prefill is compute-bound due to its highly parallel workload, whereas decode is memory-bound as it repeatedly accesses the expanding KV cache.

These contrasting resource demands lead to performance interference when both phases share the same hardware. \textbf{Prefill-Decode disaggregation} addresses this by running the two phases on separate, specialized compute clusters, enabling better resource utilization, lower tail latency, and more reliable SLO compliance \cite{dist-serve}\cite{splitwise}.

\begin{figure}[t]
  \centering
  \begin{minipage}[t]{0.48\linewidth}
    \centering
    \includegraphics[width=\textwidth]{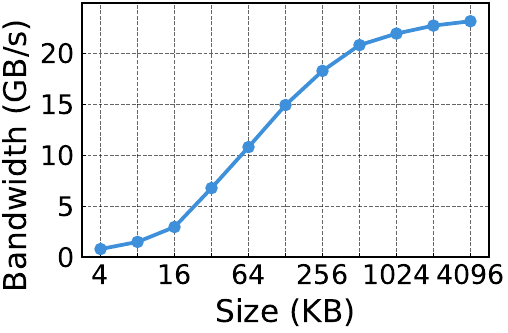}
    \caption{I/O bandwidth between GPU and CPU.}
    \label{fig:memcpy-bw}
  \end{minipage}
  \hfill
  \begin{minipage}[t]{0.48\linewidth}
    \centering
    \includegraphics[width=\textwidth]{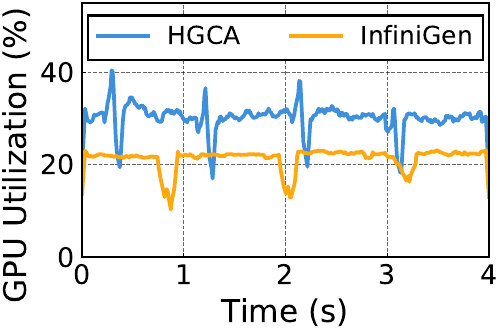}
    \caption{Low GPU utilization of HGCA and InfiniGen.}
    \label{fig:gpu-util}
  \end{minipage}
\end{figure}

\subsection{Sparse Attention}\label{sec:sparse-attn}

Attention exhibits inherent sparsity, with fewer than 20\% of tokens accounting for more than 80\% of the total attention weights. Building on this observation, a growing body of work \cite{H2O}\cite{quest}\cite{streaming-llm}\cite{snapkv} propose sparse attention algorithms that reduce computation by selectively involving only the most influential tokens in the attention computation.

Among these sparse attention algorithms, block-wise sparsity has attracted considerable interest. It segments the KV cache into fixed-size blocks and summarizes each block with a block digest $K_{\text{digest}}$. During attention computation, the top-k important blocks are selected according to the dot product between $Q$ and $K_{\text{digest}}$. Different block-wise sparse methods adopt different strategies for generating block digests. For example, Quest \cite{quest} constructs digests using channel-wise min/max pooling over $K$, MoBA \cite{moba} uses mean pooling of $K$, and NSA \cite{nsa} incorporates a learnable Multi-Layer Perceptron (MLP) to generate digests from $K$.

\subsection{Related Work and Motivation}\label{sec:motivation}

To mitigate the large GPU memory footprint associated with long-context processing, recent studies have investigated offloading the KV cache to DRAM. These approaches fall into two primary strategies: recall-based method and co-attention approach. Figure~\ref{fig:pipeline} illustrates the pipeline of different methods.

\textbf{Recall-based offloading.} Recall-based approaches involve reloading the KV cache back to GPU memory when needed for computation.
For example, InfiniGen offloads most tokens to DRAM while keeping only critical tokens on the GPU. During attention computation, it introduces a predictive mechanism that identifies the important tokens for the next layer and initiates the recall I/O one layer in advance, enabling overlap between I/O and computation.

However, interconnection bandwidth poses a critical bottleneck for these methods.
Figure~\ref{fig:memcpy-bw} demonstrates the severe I/O bandwidth constraints between a 80GB HBM GPU and CPU, which communicate through a PCIe 4x16 interface.
With a KV cache size of roughly 4 KB per token per layer, the effective I/O bandwidth is only 800 MB/s. Even when transferring at a coarser granularity using a page size of 32 tokens (128 KB), the bandwidth increases to about 15 GB/s, which remains low compared with the 1.9 TB/s HBM bandwidth.
Our evaluation reveals that despite InfiniGen's sophisticated one-layer-ahead prefetching strategy, the GPU remains idle for 61\% of execution time at a batch size of 40, resulting in very low GPU utilization, as shown in Figure~\ref{fig:gpu-util}.

\textbf{Co-attention approach.} For the co-attention approach, tokens offloaded to DRAM are computed directly on the CPU.
For instance, HGCA \cite{hgca} retains only a sliding window of 25\% tokens on the GPU, offloading the remaining 75\% tokens to the CPU. During computation, the GPU and CPU execute in parallel: the GPU-side performs full attention calculation, while the CPU-side employs sparse attention based on moving average attention weights.

Although this co-attention approach eliminates the slow I/O bottleneck present in recall-based methods, it introduces a new bottleneck caused by the significant computational disparity between the CPU and GPU. In decoding-phase attention, the GPU is roughly 20x faster than the CPU. Consequently, the parallel execution in HGCA results in about 57\% GPU idle time under batch size of 40.

\section{Design of ScoutAttention}

\begin{figure}
  \centering
  \includegraphics[width=\linewidth]{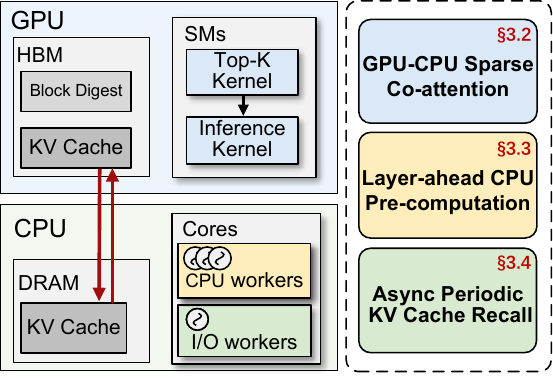}
  \caption{Overall architecture of ScoutAttention.}
  \label{fig:arch}
\end{figure}

\subsection{Overview}

We present ScoutAttention, a GPU-CPU collaborative sparse attention mechanism designed for efficient KV cache offloading.
To maximize performance, ScoutAttention is built on three key design principles: 1) \textbf{GPU-CPU co-attention}, which mitigates PCIe bandwidth bottlenecks; 2) \textbf{CPU-side pre-computation}, which hides CPU compute latency; and 3) \textbf{integration with block-wise sparse attention}, which reduces the CPU's computational burden.

Figure~\ref{fig:arch} illustrates the overall architecture of ScoutAttention. During the decoding stage, ScoutAttention offloads most unimportant KV blocks to DRAM while retaining only block digests and a small set of important blocks on the GPU. During attention computation, we employ GPU-CPU cooperative sparse attention, where the top-k blocks are processed using a near-data computing approach (\S\ref{sec:sparse-co-attn}). To prevent slow CPU-side attention computation from becoming the system bottleneck, we introduce a novel layer-ahead CPU pre-computation algorithm (\S\ref{sec:pre-compute}). In addition, to correct importance drift as decoding progresses, we propose an asynchronous periodic recall mechanism that keeps the CPU's computational load low (\S\ref{sec:periodic-recall}).

\subsection{GPU-CPU Collaborative Block-wise Sparse Attention}\label{sec:sparse-co-attn}

In modern AI servers, the CPU's attention computation throughput (KV cache size divided by compute time) is significantly higher than the KV cache transfer throughput over PCIe. A 36-core CPU can achieve an attention computation throughput of approximately 100 GB/s. However, as discussed in Sec.~\ref{sec:motivation}, the PCIe bandwidth for KV cache transfer reaches only about 15 GB/s.

Motivated by this disparity, we adapt GPU–CPU co-attention rather than KV cache recall to enable efficient KV cache offloading. Unlike HGCA, which performs the entire sparse attention computation on the CPU, our design executes cooperative block-wise sparse attention across both CPU and GPU. As demonstrated in prior works (\S~\ref{sec:sparse-attn}), block-wise sparse attention has been extensively validated for accuracy.
In this paper, we employ Quest as our sparsification method; however, ScoutAttention is fully compatible with other sparsification algorithms such as DeepSeek NSA~\cite{nsa}.

\begin{figure}
  \centering
  \includegraphics[width=\linewidth]{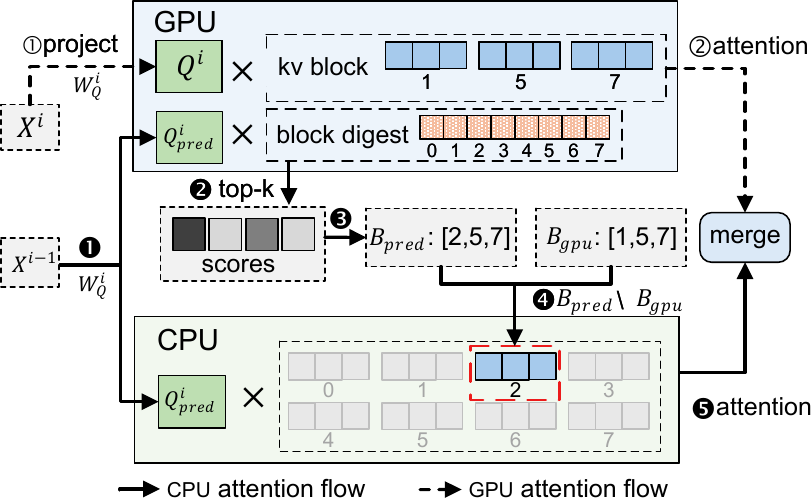}
  \vspace{1pt}
  \caption{Workflow of GPU-CPU collaborative block-wise sparse attention with layer-ahead CPU pre-computation.}
  \label{fig:attn_workflow}
\end{figure}

Building on block-wise sparse attention, ScoutAttention offloads less important KV cache blocks to DRAM while retaining only compact digests of each block and a fixed subset of critical blocks in GPU memory. During attention computation, the GPU first identifies the top-k blocks by computing the dot product between the query and each block's digest.
It then processes the blocks that reside on the GPU, while the fraction of top-k blocks that are not resident on the GPU are computed on the CPU.
Finally, the intermediate results from both devices are merged on the GPU using the Flash\-Attention algorithm \cite{flash-attention}\cite{flashattention-notes} to produce the final attention output. Figure~\ref{fig:attn_workflow} shows the overall workflow of ScoutAttention.

\begin{figure}[tb]
  \centering
  \begin{subfigure}{0.48\linewidth}
    \centering
    \includegraphics[width=\linewidth]{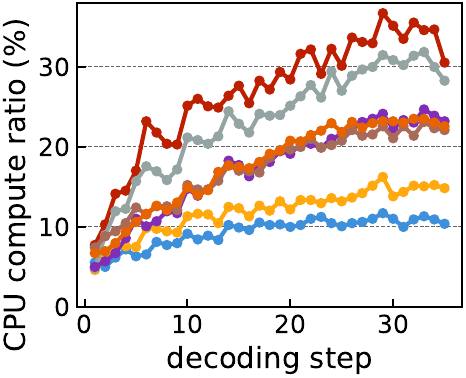}
    \caption{w/o periodic recall}
    \label{fig:wo-periodic-recall}
  \end{subfigure}
  \hfill
  \begin{subfigure}{0.48\linewidth}
    \centering
    \includegraphics[width=\linewidth]{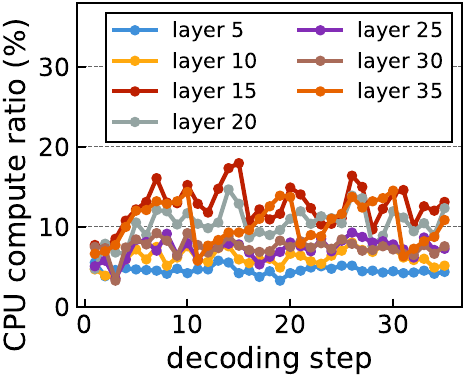}
    \caption{w/ periodic recall}
    \label{fig:w-periodic-recall}
  \end{subfigure}

  \caption{CPU compute ratio (\#token/budget) across decoding steps of Qwen 3 8B.}
  \label{fig:periodic-recall}
\end{figure}

This partitioning strategy is highly effective because important blocks exhibit strong temporal locality across adjacent tokens. As shown in Figure~\ref{fig:wo-periodic-recall}, on average less than 15\% of important blocks change between consecutive tokens.
Since the GPU retains the important blocks identified in the preceding steps, the CPU needs to process only a small fraction of top-k blocks that not resident on the GPU in ScoutAttention, substantially reducing its computational overhead.

\subsection{Layer-Ahead CPU Pre-computation}\label{sec:pre-compute}
To further prevent CPU computation from becoming a bottleneck, we propose \textbf{layer-ahead CPU pre-computation}, a novel technique that allows CPU-side attention computation to begin one layer ahead of the GPU's execution.

As illustrated in Figure~\ref{fig:pipeline}, ScoutAttention employs a pipelined execution strategy. While computing attention for layer i on the GPU, ScoutAttention first identifies the top-k important blocks for layer i+1. This enables proactive triggering of CPU-side attention computation for layer i+1, allowing it to run in parallel with the GPU's processing of layer i. The GPU then computes the attention for layer i and merges it with the CPU-side attention results that were triggered during layer i-1.

To enable CPU-side pre-computation, the key challenge is determining how to obtain layer i+1's attention query at layer i. Inspired by InfiniGen, which observes that the inputs of consecutive layers are highly similar due to residual connections, we hypothesize that it is possible to approximate $Q^{i+1}$ by applying the next layer's query projection matrix $W_Q^{i+1}$ to the current layer's input $X^{i}$, yielding a predicted query $Q_{\text{pred}}^{i+1}$.
To validate this hypothesis, we conducted experiments across various models and observed that the cosine similarity between $Q_{\text{pred}}^{i+1}$ and $Q^{i+1}$ remains consistently high, as shown in Table~\ref{tab:cos-sim}. This confirms that the predicted query provides a reliable approximation for CPU-side pre-computation.

\begin{algorithm}[t]
\caption{Layer-ahead CPU Pre-computation}
\label{alg:cpu-precomputation}
\begin{spacing}{1.25}
\begin{algorithmic}[1]
\State \textbf{Input:} layer i's input $X^{i}$, query projection weight $W_Q$, block digest $K_\text{digest}$.
\State \textbf{Output:} layer i's attention output $A^{i}$.

\LComment{Trigger next layer's CPU-side pre-computation}

\State $Q_{\text{pred}}^{i+1} \gets W_Q^{i+1} X^{i}$ \Comment{Get predict query of next layer}
\State $B_{\text{pred}}^{i+1} \gets \Call{TopK}{Q_{\text{pred}}^{i+1} K_{\text{digest}}^{i+1}{}^{\top}$} \Comment{Predict top-k blocks}
\State $B_{\text{cpu}}^{i+1} \gets B_{\text{pred}}^{i+1} \setminus B_{\text{gpu}}^{i+1}$ \Comment{Blocks reside in CPU}
\State \textbf{spawn} \Call{CPUAttn}{$B_{\text{cpu}}^{i+1}$} \Comment{Async CPU pre-computation}

\LComment{Compute current layer's attention}
\State $Q^{i} \gets W_Q^{i} X^{i}$
\State $A_{\text{gpu}}^{i} \gets \Call{GPUAttn}{Q^{i},\, B_{\text{gpu}}^{i}}$ 
\LComment{Merge with $A_{\text{cpu}}^{i}$ triggered in previous layer}
\State $A^{i} \gets \Call{Merge}{A_{\text{gpu}}^{i},\, A_{\text{cpu}}^{i}}$

\State \Return $A^{i}$
\end{algorithmic}
\end{spacing}
\end{algorithm}

\begin{table}[tb]
\centering
\caption{Cosine similarity between predict query and real query.}
\label{tab:cos-sim}
\begin{tabular}{ccccc}
\toprule
\textbf{\makecell{Qwen 3\\8B}} & \textbf{\makecell{Gemma 3\\12B}} & \textbf{\makecell{Llama 3.1\\8B}} & \textbf{\makecell{Mistral\\7B}} & \textbf{\makecell{GLM 4\\9B}} \\
\midrule
0.94 & 0.93 & 0.96 & 0.97 & 0.94 \\
\bottomrule
\end{tabular}
\end{table}

Building on this, we design a pre-computation algorithm as described in Algorithm~\ref{alg:cpu-precomputation} and illustrated in Figure~\ref{fig:attn_workflow}.
When the GPU is computing layer i, it first predicts the next layer's query representation $Q_{\text{pred}}^{i+1}$ by applying the projection matrix $W_Q^{i+1}$ to the current layer's input $X^{i}$. Using $Q_{\text{pred}}^{i+1}$ and the next layer's digest key $K_{\text{digest}}^{i+1}$, the model then identifies the important blocks of the next layer by computing the dot product between them and selecting the top-k blocks, denoted as $B_{\text{pred}}^{i+1}$.
Next, we compare $B_{\text{pred}}^{i+1}$ with the blocks already resident in GPU memory $B_{\text{gpu}}^{i+1}$. The blocks that are not yet on the GPU are marked as $B_{\text{cpu}}^{i+1}$. These blocks are then triggered to perform pre-computation on the CPU.
After completing this setup of layer i+1's CPU-side pre-computation, the GPU proceeds with layer i's attention computation. It first performs the GPU-side attention to produce $A_{\text{gpu}}^{i}$, and then merges this result with the CPU-side output $A_{\text{cpu}}^{i}$, which was pre-computed during the previous layer, to obtain the final attention output $A^{i}$.

Instead of executing CPU operations concurrently with GPU computation, our pre-computation strategy leverages the entire transformer layer's processing window for CPU-side attention computation. This extended window spans GPU-side attention, feed-forward networks, and QKV projections. On an 80G HBM GPU running Qwen3-32B with a 4k sparsification budget, attention computation alone requires 300us, while the complete transformer layer consumes 900us during decoding. This layer-ahead pre-computation thus grants the CPU 3x more processing time compared to parallel computation approaches, effectively eliminating GPU idle periods.

\subsection{Asynchronous Periodic KV Cache Recall}\label{sec:periodic-recall}

As decoding progresses, the set of important KV cache blocks gradually shifts. Because the CPU handles the KV blocks that are not resident on the GPU, whereas the GPU retains the important blocks identified after the prefill phase, this shift causes the CPU to process an increasingly large portion of tokens. Figure~\ref{fig:wo-periodic-recall} illustrates the CPU compute ratio, defined as the number of tokens computed on the CPU relative to the sparse budget size at each decoding step. The ratio is low in the early decoding steps but shows a steady upward trend as generation proceeds, reflecting the drift in block importance.

To mitigate performance degradation, we propose an asynchronous periodic KV cache recall mechanism. This process refreshes GPU-resident context by strategically recalling important blocks from DRAM at regular intervals. We recognize that the PCIe interconnection is a significant bottleneck for fine-grained KV cache transfers. Therefore, our design is centered on two principles: minimizing recall frequency and moving the data transfer operation off the critical inference path. Specifically, if the system decides to trigger a recall at layer i of decoding step m, ScoutAttention initiates the I/O operation \emph{after} the attention computation for layer i is complete. This asynchronous approach provides a sufficient time window for the transfer, as the recalled blocks are not required by the GPU until layer i of the next decoding step m+1---typically offering a window exceeding 20ms.

The recall interval is determined empirically via offline profiling. For this, we track the CPU compute ratio across decoding steps as in Figure~\ref{fig:wo-periodic-recall}. We observe that different layers exhibit distinct patterns, leading us to establish per-layer recall intervals.
For each layer, we determine the maximum number of steps that keeps the measured ratio below a threshold $\beta$.
This threshold, determined by the system's GPU/CPU computational power ratio and interconnection bandwidth, balances CPU computation overhead against I/O volume.
Based on our empirical analysis, we set the default threshold at 12\%. Figure~\ref{fig:w-periodic-recall} illustrates the improved CPU compute ratio achieved through our asynchronous periodic KV cache recall mechanism, where the average CPU compute ratio is 8.2\% and the average recall interval across all layers is 8.7.

\section{Implementation and Evaluation}

We implement ScoutAttention on SGLang \cite{sglang}, one of the most widely used inference frameworks.
To efficiently identify the top-k blocks, we implement a block-wise top-k selection CUDA kernel based on FlashInfer \cite{flashinfer}. For CPU-side attention computation, we build an optimized CPU attention worker using IPEX (Intel Extension for PyTorch) \cite{intel-ipex}. We further partition CPU threads into groups, with each group handling one sequence in the batch, thereby improving overall attention computation throughput.

\begin{figure}[tb]
  \centering
  \includegraphics[width=\linewidth]{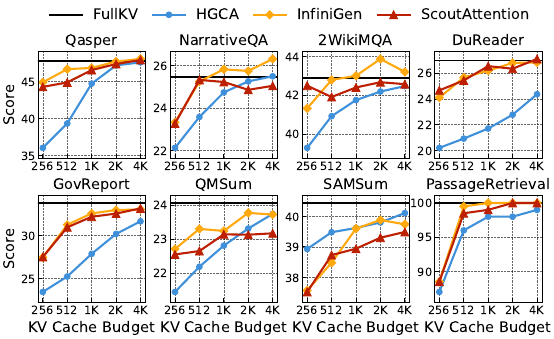}
  \caption{Results on LongBench of different methods.}
  \label{fig:accuracy}
\end{figure}

\subsection{Baselines}

We select three baselines: FullKV, which corresponds to a vanilla Transformer; InfiniGen, a recall-based KV cache offloading method; and HGCA, a co-attention-based KV cache offloading approach. All baselines are implemented in SGLang to ensure a fair comparison.

\subsection{Accuracy Benchmark}

We use LongBench \cite{longbench}, a widely adopted long context understanding benchmark for LLMs, to evaluate the accuracy of ScoutAttention. We select eight datasets, including Qasper, NarrativeQA, 2WikiMQA, DuReader, GovReport, QMSum, SAMSum, and PassageRetrieval. These datasets cover a variety of tasks including single document QA, multi document QA, summarization, and message retrieval. The context lengths reach up to 64k tokens.

We evaluate ScoutAttention and baselines across different budget sizes, with all experiments conducted using the Qwen 3 8B model.
As shown in Figure~\ref{fig:accuracy}, ScoutAttention maintains robust accuracy across datasets. The slight accuracy drop relative to InfiniGen comes from using predicted queries for CPU side attention computation. However, because the cosine similarity between predicted and real queries is very high (see Table~\ref{tab:cos-sim}) and CPU-side computation accounts for only 8\% of the total budget (\S\ref{sec:periodic-recall}), replacing real queries with predicted ones introduces minimal accuracy loss. Compared with full attention, ScoutAttention's accuracy decreases by only 2.5\% on average at a budget size of 1024 and 2.1\% at a budget size of 2048.

\subsection{Performance Evaluation}

\begin{figure}[tb]
  \centering
  \includegraphics[width=\linewidth]{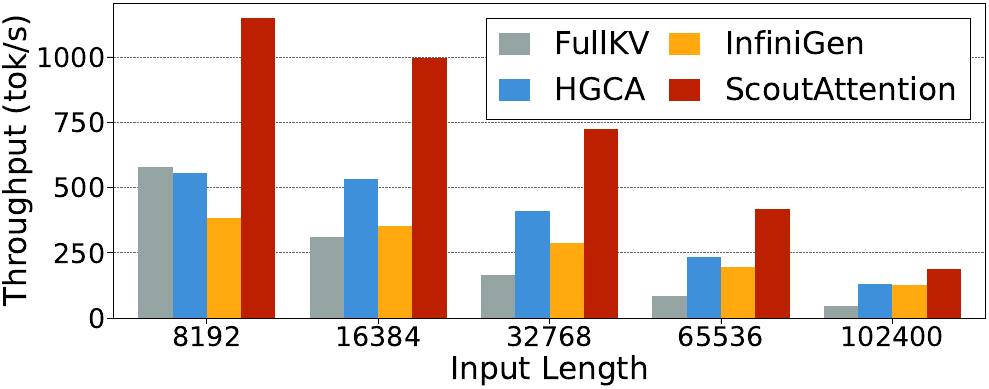}
  \caption{Decode throughput of different methods under varying input length.}
  \label{fig:input_len_to_throughput}
\end{figure}

In performance evaluations, we focus solely on the decode instances in Prefill–Decode disaggregation architectures, evaluating only the decode phase while the prefill phase is handled separately.
For all baseline methods, the sparse budget is fixed at 2,048 tokens, and the block size in ScoutAttention is set to 32. All experiments in this section are conducted using the Qwen 3 14B model.

\textbf{Overall throughput.} We first evaluate decoding throughput under varying input lengths.
As shown in Figure~\ref{fig:input_len_to_throughput}, ScoutAttention consistently achieves the highest throughput across different sequence lengths.
Moreover, as the input length increases, the speedup over FullKV grows steadily due to FullKV's severe memory-capacity bottleneck. At an input length of 64k, ScoutAttention attains a 5.1x speedup over FullKV.
For HGCA and InfiniGen, the large GPU idle time caused by I/O bottlenecks and CPU compute bottlenecks (\S\ref{sec:motivation}) results in throughput that is even lower than FullKV at an 8k input length. Although both methods eventually surpass FullKV at larger input lengths, ScoutAttention still delivers up to a 2.1x speedup compared to them.

\begin{figure}[t]
  \centering
  \begin{minipage}[t]{0.51\linewidth}
    \centering
    \includegraphics[width=\textwidth]{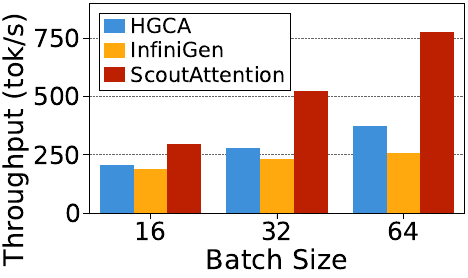}
    \caption{Decode throughput under different batch size.}
    \label{fig:batch_size_to_throughput}
  \end{minipage}
  \hfill
  \begin{minipage}[t]{0.46\linewidth}
    \centering
    \includegraphics[width=\textwidth]{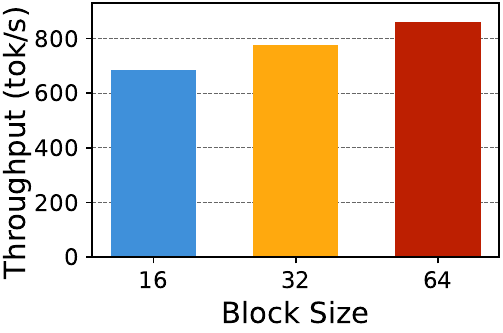}
    \caption{Decode throughput under varying block size.}
    \label{fig:block_size_to_throughput}
  \end{minipage}
\end{figure}

\textbf{Impact of batch size and block size.}
Figure~\ref{fig:batch_size_to_throughput} illustrates throughput scaling across batch sizes for a 32k input length.
HGCA and InfiniGen exhibit sublinear scaling and achieve only 1.31x and 1.21x speedups when increasing the batch size from 16 to 32, as they are constrained by CPU compute and I/O bottlenecks.
In contrast, ScoutAttention demonstrates better scalability, achieving a 1.78x speedup from batch 16 to 32 and a 1.48x speedup from batch 32 to 64.

Figure \ref{fig:block_size_to_throughput} illustrates the throughput of ScoutAttention on three different block sizes. As the block size increases, the total number of blocks decreases. This reduces the size of the digest cache used for block selection, freeing up memory to support larger batch sizes and enhancing decode throughput. 

\begin{figure}[t]
  \centering
  \begin{minipage}[t]{0.51\linewidth}
    \centering
    \includegraphics[width=\textwidth]{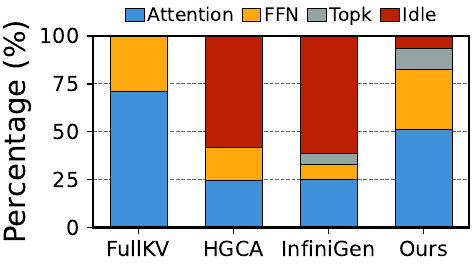}
    \caption{Latency Breakdown.}
    \label{fig:latency_breakdown}
  \end{minipage}
  \hfill
  \begin{minipage}[t]{0.46\linewidth}
    \centering
    \includegraphics[width=\textwidth]{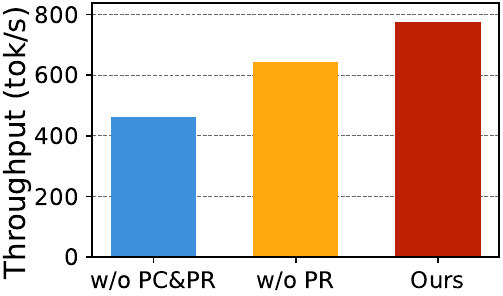}
    \caption{Ablation study.}
    \label{fig:ablation}
  \end{minipage}
\end{figure}

\textbf{Latency breakdown.}
Figure~\ref{fig:latency_breakdown} presents the end-to-end latency breakdown, where \emph{idle} represents GPU stalls due to CPU computation dependencies or PCIe data transfers. HGCA experiences severe CPU compute bottlenecks, with idle time comprising 57\% of total latency. InfiniGen performs even worse, with idle time rising to 61\% due to significant I/O bottlenecks. In contrast, ScoutAttention dramatically reduces idle time to merely 6\% by minimizing CPU computation and employing CPU-side pre-computation.

\textbf{Ablation study.}
We perform an ablation study on ScoutAttention to quantify the performance contribution of each optimization. As shown in Figure~\ref{fig:ablation}, ``PC'' refers to pre-computation and ``PR'' to periodic recall.
By hiding CPU compute latency through layer-ahead pre-computation, PC achieves a 1.39x speedup. Meanwhile, asynchronous periodic KV-cache recall provides an additional 1.20x speedup by reducing the CPU compute load.

\section{Conclusion}

We propose ScoutAttention, an efficient GPU–CPU co-attention mechanism for KV cache offloading.
To enable high-performance CPU-side computation, we introduce a novel layer-ahead CPU pre-computation algorithm paired with an asynchronous periodic recall mechanism. Experimental results show that ScoutAttention achieves only a 2.1\% accuracy drop while delivering up to a 5.1x speedup compared to full attention.

\begin{acks}
This work was sponsored by the National Key Research and Development Program of China under Grant No.2023YFB4502701, the National Natural Science Foundation of China under Grant No.62502170 and No.62302465, the China Postdoctoral Science Foundation under Grant No.2024M751011, and the Postdoctor Project of Hubei Province under Grant No.2024HBBHCXA027.
\end{acks}

\clearpage
\bibliographystyle{ACM-Reference-Format}
\bibliography{reference}

\end{document}